%% file: main.tex
\documentclass[sigconf]{acmart}

\AtBeginDocument{%
  \providecommand\BibTeX{{%
    \normalfont B\kern-0.5em{\scshape i\kern-0.25em b}\kern-0.8em\TeX}}}

\copyrightyear{2024} 
\acmYear{2024} 
\setcopyright{rightsretained} 
\acmConference[KDD '24]{Proceedings of the 30th ACM SIGKDD Conference on Knowledge Discovery and Data Mining}{August 25--29, 2024}{Barcelona, Spain}
\acmBooktitle{Proceedings of the 30th ACM SIGKDD Conference on Knowledge Discovery and Data Mining (KDD '24), August 25--29, 2024, Barcelona, Spain}
\acmDOI{10.1145/3637528.3671975}
\acmISBN{979-8-4007-0490-1/24/08}

\usepackage{xcolor}

\usepackage{multirow}
\usepackage{longtable}
\usepackage{{booktabs}}
\usepackage{amsmath}
\usepackage{graphicx}
\usepackage{paralist}

\input{KDD2024_Camera/math_commands}

\usepackage{subcaption}
\usepackage{overpic}  

\usepackage{hyperref}
\usepackage{url}
\usepackage{listings}
\usepackage{color}
\definecolor{jsonString}{rgb}{0.08,0.08,0.80}
\definecolor{jsonNumber}{rgb}{0.08,0.64,0.08}
\definecolor{jsonBoolean}{rgb}{0.31,0.60,0.02}
\definecolor{jsonNull}{rgb}{0.31,0.60,0.02}
\definecolor{jsonPunctuation}{rgb}{0.31,0.60,0.02}
\lstdefinelanguage{json}{  
    basicstyle=\normalfont\ttfamily,  
    numbers=left,  
    numberstyle=\scriptsize,  
    stepnumber=1,  
    numbersep=8pt,  
    showstringspaces=false,  
    breaklines=true,  
    frame=lines,  
    backgroundcolor=\color{white},  
    literate=  
     *{0}{{{\color{jsonNumber}0}}}{1}  
      {1}{{{\color{jsonNumber}1}}}{1}  
      {2}{{{\color{jsonNumber}2}}}{1}  
      {3}{{{\color{jsonNumber}3}}}{1}  
      {4}{{{\color{jsonNumber}4}}}{1}  
      {5}{{{\color{jsonNumber}5}}}{1}  
      {6}{{{\color{jsonNumber}6}}}{1}  
      {7}{{{\color{jsonNumber}7}}}{1}  
      {8}{{{\color{jsonNumber}8}}}{1}  
      {9}{{{\color{jsonNumber}9}}}{1}  
      {:}{{{\color{jsonPunctuation}{:}}}}{1}  
      {,}{{{\color{jsonPunctuation}{,}}}}{1}  
      {\{}{{{\color{jsonPunctuation}{\{}}}}{1}  
      {\}}{{{\color{jsonPunctuation}{\}}}}}{1}  
      {[]}{{{\color{jsonPunctuation}{[]}}}}{1}  
      {true}{{{\color{jsonBoolean}true}}}{1}  
      {false}{{{\color{jsonBoolean}false}}}{1}  
      {null}{{{\color{jsonNull}null}}}{1}  
      {=}{{{\color{jsonPunctuation}==}}}{1},  
    stringstyle=\color{jsonString},  
    morestring=[b]",  
} 
\begin{document}

\title{From Supervised to Generative: A Novel Paradigm for Tabular Deep Learning with Large Language Models}

\author{Xumeng Wen}
\affiliation{%
  \institution{Microsoft Research Asia}
  \city{Beijing}
  \country{China}}
\email{xumengwen@microsoft.com}

\author{Han Zhang}
\authornote{Han did this work during his internship at Microsoft Research Asia, Beijing, China.}
\affiliation{%
  \institution{Tsinghua University}
  \city{Beijing}
  \country{China}}
\email{zh950713@gmail.com}

\author{Shun Zheng}
\affiliation{%
  \institution{Microsoft Research Asia}
  \city{Beijing}
  \country{China}}
\email{shun.zheng@microsoft.com}

\author{Wei Xu}
\affiliation{%
  \institution{Tsinghua University}
  \city{Beijing}
  \country{China}}
\email{weixu@tsinghua.edu.cn}

\author{Jiang Bian}
\affiliation{%
  \institution{Microsoft Research Asia}
  \city{Beijing}
  \country{China}}
\email{jiang.bian@microsoft.com}

\renewcommand{\shortauthors}{Xumeng Wen, Han Zhang, Shun Zheng, Wei Xu, and Jiang Bian}

\begin{abstract}
Tabular data is foundational to predictive modeling in various crucial industries, including healthcare, finance, retail, sustainability, etc.
Despite the progress made in specialized models, there is an increasing demand for universal models that can transfer knowledge, generalize from limited data, and follow human instructions. These are challenges that current tabular deep learning approaches have not fully tackled.
Here we introduce Generative Tabular Learning (GTL), a novel framework that integrates the advanced functionalities of large language models (LLMs)—such as prompt-based zero-shot generalization and in-context learning—into tabular deep learning. GTL capitalizes on the pre-training of LLMs on diverse tabular data, enhancing their understanding of domain-specific knowledge, numerical sequences, and statistical dependencies critical for accurate predictions.
Our empirical study spans 384 public datasets, rigorously analyzing GTL's convergence and scaling behaviors and assessing the impact of varied data templates. The GTL-enhanced LLaMA-2 model demonstrates superior zero-shot and in-context learning capabilities across numerous classification and regression tasks. Notably, it achieves this without fine-tuning, outperforming traditional methods and rivaling state-of-the-art models like GPT-4 in certain cases.
Through GTL, we not only foster a deeper integration of LLMs' sophisticated abilities into tabular data comprehension and application but also offer a new training resource and a test bed for LLMs to enhance their ability to comprehend tabular data.
To facilitate reproducible research, we release our code, data, and model checkpoints at \url{https://github.com/microsoft/Industrial-Foundation-Models}.
\end{abstract}

\ccsdesc[500]{Computing methodologies~Neural networks}
\ccsdesc[500]{Computing methodologies~Modeling methodologies}
\ccsdesc[300]{Computing methodologies~Transfer learning}

\keywords{tabular data, large language models, generative modeling, instruction following, in-context learning, zero-shot learning}



\maketitle

\input{KDD2024_Camera/1_introduction}
\input{KDD2024_Camera/2_related_work}
\input{KDD2024_Camera/3_methodology}

\input{KDD2024_Camera/4_experiments}
\input{KDD2024_Camera/5_conclusion}

\bibliographystyle{KDD2024_Camera/ACM-Reference-Format}
\bibliography{my_ref}

\appendix
\input{KDD2024_Camera//6_appendix}

\end{document}

%% file: KDD2024_Camera/math_commands.tex
\usepackage{amsmath,amsfonts,bm}

\def\eqref#1{equation~\ref{#1}}

\def\1{\bm{1}}

\DeclareMathAlphabet{\mathsfit}{\encodingdefault}{\sfdefault}{m}{sl}
\SetMathAlphabet{\mathsfit}{bold}{\encodingdefault}{\sfdefault}{bx}{n}

\newcommand{\R}{\mathbb{R}}



%% file: KDD2024_Camera/1_introduction.tex
\section{Introduction}
\label{sec:intro}

Tabular data, with its widespread presence across numerous critical industrial domains such as healthcare, finance, retail, sustainability, and climate~\citep{provost2013ds4business,chen2016XGBoost,shwartzziv2022tab_dl_is_not_all,hegselmann2023TabLLM}, serves as a cornerstone for predictive modeling~\citep{kelleher2020fundamentalsML}.
This modeling underpins a diverse array of real-world applications, including disease risk stratification, credit assessment, sales volume prediction, grid stability measurement, climate estimation, etc.
Given its significance, it has attracted substantial research attention from the machine learning community.


While there have been considerable advancements in developing specialized predictive models for individual tasks within tabular data, employing effective learning models such as tree ensemble models~\citep{chen2016XGBoost,prokhorenkova2018catboost,ke2017LightGBM} based on gradient boosting decision trees~\cite{friedman2001GBDT} and more recent neural networks~\citep{huang2020TabTransformer,arik2021TabNet,katzir2021Net-DNF,gorishniy2021revisit_tab_dnn}, the rich and diverse application scenarios across various domains have underscored a critical need for universal models. These models should possess the ability to seamlessly transfer to new datasets and exhibit robust generalization capabilities, especially in scenarios with few-shot labels. This necessity has catalyzed the emergence of tabular deep learning~\citep{gorishniy2021revisit_tab_dnn} as a research focus, which explores the concept of pre-training a universal neural network on a broad spectrum of tabular data. This approach aims to create models that, once pre-trained, can be effortlessly adapted to a wide range of tasks, thereby enhancing their utility and efficiency across different applications~\citep{yoon2020VIME,ucar2021SubTab,somepalli2021SAINT,bahri2022SCARF,nam2023STUNT,wang2022TransTab,hollmann2022TabPFN,levin2023transfer_tab_nn,zhu2023XTab,yang2023UniTabE}.



The rapidly evolving foundation models in language and multimodal domains have illuminated the potential for pre-training universal deep models for tabular data, showcasing remarkable generalization capabilities~\citep{Brown2020GPT-3,OpenAI2023GPT4TR,huang2023cosmos1,google2023Gemini}. 
Specifically, these foundation models highlight two key attributes: Firstly, upon pre-training, they can rapidly adapt to new tasks simply by specifying a task-oriented prompt~\citep{ouyang2022instru_follow}. This form of adaptation is not only user-friendly but also critically enhances the model's ability to transfer knowledge acquired during pre-training to novel, low-resource scenarios efficiently. 
Secondly, in situations involving few-shot labels, these models exhibit an intriguing in-context learning ability, evidenced by their impressive capacity to learn from demonstrations included as part of the prompt~\citep{Brown2020GPT-3}. This capability, without fine-tuning, implicitly leverages Bayesian inference~\citep{xie2022explainICL} to extend the model's pre-trained knowledge to new data samples directly during forward propagation. 
Together, these attributes suggest a promising direction for advancing tabular deep learning by integrating the adaptability and learning efficiencies found in foundation models.


But most existing pre-training methods for tabular data predominantly adhere to the supervised paradigm, necessitating a subsequent fine-tuning process for adaptation to new tabular tasks. 
Despite pioneering efforts to transcend these limitations through advanced generalization capabilities, i.e., zero-shot learning to unseen tasks~\citep{hegselmann2023TabLLM,yang2023UniTabE,wang2022TransTab} and in-context learning based on fresh tabular instances~\citep{dinh2022LIFT,hollmann2022TabPFN}, the outcomes often encapsulate constrained generalization abilities and a limited scope of application.
For example, LIFT~\citep{dinh2022LIFT}, comparing fine-tuning with in-context learning on six classifications, uses a narrow evaluation scope by concentrating exclusively on accuracy, which may not capture the model's comprehensive generalization capability and its efficacy in diverse scenarios.
Similarly, TabPFN~\citep{hollmann2022TabPFN}, which develops a prior-data fitted network~\citep{muller2021TransDoBayInf} and primarily caters to classification datasets with numerical features, exhibits a limitation by not accommodating zero-shot inference.
Additionally, while TabLLM~\citep{hegselmann2023TabLLM}, UniTabE~\citep{yang2023UniTabE}, and TransTab~\citep{wang2022TransTab} have showcased successful zero-shot demonstrations, primarily in a handful of classification tasks, they often neglected in-context learning. Instead, they depend on fine-tuning for adapting models to downstream tasks, missing opportunities for more nuanced data-driven adaptation.
The limited scope of these models underscores the demand for a more holistic tabular deep learning paradigm that not only embraces but also extends the advanced generalization capabilities found in other cutting-edge foundation models.



Nonetheless, integrating the sophisticated capabilities of modern large language models (LLMs)~\citep{Brown2020GPT-3,OpenAI2023GPT4TR} - specifically, prompt-based zero-shot generalization and in-context learning — into tabular deep learning presents significant challenges. A primary issue is that LLMs, primarily pre-trained on language data, often struggle to fully grasp the nuances of language-formatted tabular data. Despite their proficiency in acquiring broad world knowledge and advanced reasoning skills from textual corpora, these models frequently lack the ability to understand domain-specific knowledge that is crucial for effective tabular deep learning. This shortfall is particularly evident in their handling of numerical features represented as token sequences and in identifying the intricate statistical dependencies that exist between prediction targets and tabular features, as well as those between in-context demonstrations and data samples for prediction. These competencies are essential for comprehending the unique characteristics of tabular datasets and leveraging them for accurate predictive modeling.

To address this shortcoming, we introduce \emph{Generative Tabular Learning} (GTL) - a novel paradigm that advocates for continued pre-training of an LLM on extensive tabular data, transcribed in an instruction-oriented language format, spanning multiple domains. GTL is meticulously designed to foster an enhanced comprehension of tabular features, their interrelations with prediction targets and in-context data examples, and the linkage between task instructions and tabular predictions. To facilitate this process, we create specific text templates to convert a tabular data instance into an instruction-oriented language format. This conversion caters to various configurations, such as including meta-information and in-context examples or not, and maintains a record of the positions of numerical tokens and target tokens. 
As a result, GTL equips the model with advanced instruction-following abilities tailored for tabular deep learning, enabling it to generatively adapt to downstream tasks by interpreting natural language prompts for new task instructions or in-context examples.

In our experiments, we have compiled $384$ public tabular datasets, covering $176$ classification and $208$ regression tasks across a variety of industrial domains. For holdout evaluation, we randomly selected $20$ classification and $24$ regression tasks, while all other tasks were utilized for GTL. Besides, we employed LLaMA-2~\cite{touvron2023llama2} as our base LLM.
Our experimental investigation includes a study of GTL's convergence behaviors, demonstrating effective knowledge transfer from learned datasets to unseen ones across dataset domains. Furthermore, our exploration of GTL's scaling laws reveals key factors that significantly influence the model's generalization capability. Our investigation also encompasses an examination of the performance disparities when employing different data templates to convert tabular data instances into pure text, enabling us to analyze their advantages and drawbacks in various scenarios.

After optimizing the GTL configuration on LLaMA-2, we compared the resulting LLaMA-GTL model with competitive baselines.
In stark contrast to traditional tabular models, which require fine-tuning, LLaMA-GTL, utilizing only in-context inference without any parameter updates, achieves state-of-the-art performance in numerous extremely few-shot cases ($\le$ 32 data samples).
Even when the number of shots increases, displaying robust statistical patterns, LLaMA-GTL maintains a performance level comparable to the best performing baseline.
Additionally, TabPFN, a representative tabular baseline with in-context learning, currently only supports classification tasks.
However, LLaMA-GTL extends its coverage to regression tasks, achieving even greater performance improvements over existing fine-tuning baselines in this area.
Further, we compared LLaMA-GTL with GPT-4~\citep{OpenAI2023GPT4TR}, perhaps the most advanced LLM to date.
Despite the comparison being somewhat disadvantageous for LLaMA-GTL in terms of model size and potential data contamination risk in GPT-4~\citep{bordt2023elephants}, LLaMA-GTL has surpassed GPT-4 in many classification tasks and significantly narrowed the gap between LLaMA and this proprietary LLM in regression tasks.

Our contributions can be summarized as follows:

\begin{compactitem}
\item 
We propose GTL, a generative learning paradigm that extends the advanced capabilities of LLMs—specifically, prompt-based zero-shot generalization and in-context learning—to the realm of tabular deep learning. GTL uniquely enhances the model's understanding of tabular data by continuing the pre-training process on a diverse array of tabular datasets formatted in an instruction-oriented language, addressing the critical gap in current methodologies.
\item 
To support GTL, we have developed a comprehensive data construction pipeline that transforms tabular instances into an instruction-oriented language format. This pipeline not only fosters further research into instruction-following capabilities within tabular deep learning but also enriches the training and evaluation resources available for LLMs, aiming to bolster their comprehension and predictive accuracy with tabular data.
\item 
To demonstrate the efficacy of GTL, we have trained the LLaMA-GTL model, exhibiting its exceptional zero-shot and in-context learning performance across a variety of classification and regression tasks on tabular data. LLaMA-GTL has set new benchmarks in adaptability and generalization for tabular models, surpassing traditional methods and even state-of-the-art LLMs like GPT-4 in specific tasks.
\end{compactitem}

%% file: KDD2024_Camera/2_related_work.tex
\section{Related Work}
\label{sec:related_work}

Our literature review encompasses four aspects.
\paragraph{Predictive Modeling on Tabular Data}
The development of effective algorithms for predictive modeling on tabular data has been a longstanding research topic. In the early days, tree-based models were found to be particularly suitable for tabular data, leading to the development of several gradient boosting decision trees~\citep{chen2016XGBoost,ke2017LightGBM,prokhorenkova2018catboost}. Subsequently, as deep learning gained prominence~\citep{lecun2015DL}, numerous studies attempted to create suitable network architectures for tabular data~\citep{huang2020TabTransformer,arik2021TabNet,katzir2021Net-DNF,gorishniy2021revisit_tab_dnn} and introduced self-supervised learning schemes~\citep{yoon2020VIME,somepalli2021SAINT,ucar2021SubTab,bahri2022SCARF}. Despite these tabular neural networks not consistently outperforming tree-based models~\citep{grinsztajn2022tree_gt_tab_nn,shwartzziv2022tab_dl_is_not_all,gorishniy2021revisit_tab_dnn}, further research has advanced tabular deep learning and established new state-of-the-art results in few-shot~\citep{nam2023STUNT,hollmann2022TabPFN} and transfer learning scenarios~\citep{wang2022TransTab,levin2023transfer_tab_nn,zhu2023XTab}. However, these techniques typically required re-training or fine-tuning to adapt to new data schemas and tasks. Moreover, as mentioned in the introduction, while some advancements have introduced support for either zero-shot or in-context learning capabilities, they usually address limited scenarios. In contrast, this paper embarks on a comprehensive exploration of zero-shot and in-context generalization within the sphere of tabular deep learning. Our focus lies in the ease of adaptation without the necessity for parameter tuning, extreme data efficiency, and wide-ranging transferability across diverse domain knowledge, data schemas, and tasks. These are areas that have not been sufficiently explored in the existing literature on tabular deep learning.

\paragraph{When LLMs meet Tabular Deep Learning}
The remarkable success of LLMs, scaled to unprecedented sizes and trained on massive text corpus, has demonstrated their broad knowledge and phenomenal capabilities in transfer learning and instruction following~\citep{Brown2020GPT-3,ouyang2022instru_follow}. This success has spurred the application of LLMs to tabular deep learning with the aims of  1) leveraging the comprehensive world knowledge already acquired, 2) enabling instruction following to support diverse tasks without the need for tuning or even data, and 3) effectively harnessing meta-information in tabular data, such as column names, task descriptions, and background knowledge.
In this domain, several pioneering studies have set the stage.
For instance, LIFT~\citep{dinh2022LIFT} introduced language-interfaced fine-tuning, which fine-tuned GPT-3~\citep{Brown2020GPT-3} and GPT-J~\citep{wang2021GPT-J} on multiple tabular learning datasets, revealing that the performance of fine-tuned LLMs was roughly on par with traditional solutions. Specifically, it examined the comparison between fine-tuning and in-context learning, albeit only on six classification tasks using the accuracy metric.
TabLLM~\citep{hegselmann2023TabLLM}, a subsequent study that adopted T0~\citep{sanh2022T0} as the base LLM, demonstrated competitive performance of fine-tuned LLMs in very few-shot scenarios, and it slightly underperformed in comparison to classical tabular models when more shots were available. TabLLM explored zero-shot learning on tabular data, but it only covered classification tasks, its inference design necessitated multiple forward passes over the base LLM for multi-class tasks, and it did not explore in-context learning.
Moreover, MediTab~\citep{wang2023AnyPredict}, which focused on the healthcare domain, utilized LLMs to generate supplementary data for a specific target task, and TapTap~\citep{zhang2023TapTap} employed LLMs to create synthetic tabular data.
Additionally, we have recently identified two contemporaneous studies, TP-BERTa~\citep{anonymous2024TP-BERTa} and UniPredict~\citep{wang2024unipredict}, which also investigated the pre-training of LLMs on tabular data. However, these studies still conformed to the supervised learning paradigm for downstream tasks.
Distinct from these explorations merging LLMs with tabular deep learning, our research accentuates a generative learning paradigm for LLMs on tabular data, promoting comprehensive instruction-following capabilities for both zero-shot and in-context learning. Furthermore, our approach supports not only binary classification tasks but also multi-class and regression cases across various domains.

\paragraph{Augmented LLMs}
Since the groundbreaking success of LLMs~\citep{Brown2020GPT-3}, ongoing efforts have been made to augment LLMs with capabilities that are difficult to acquire through next-token predictions on pure text alone~\citep{mialon2023aug_lms}. These efforts can generally be divided into two categories. The first category leverages external sources or tools to gather additional information, thereby endowing LLMs with unprecedented capabilities. Notable examples in this category include ReAct~\citep{yao2023ReAct}, which augments LLMs with a simple Wikipedia API, PAL~\citep{gao2023PAL}, which combines LLMs with Python interpreters, and Toolformer~\citep{schick2023toolformer}, which teaches LLMs to use multiple tools. The second category introduces an additional learning procedure on new data, thereby inherently acquiring some new abilities. For instance, ~\cite{ouyang2022instru_follow} further trained LLMs to align with human values, and ~\cite{chen2021GPT-3-Codex} trained LLMs on code data, facilitating remarkable code understanding and generation. Viewed from the lens of augmented LLMs, this paper aligns with the second category. Our distinct contributions encompass the introduction of tabular data as a novel learning resource and testing platform for LLMs, along with the formulation of effective learning objectives for this data type.



%% file: KDD2024_Camera/3_methodology.tex
\begin{figure*}[!tb]
    \centering
    \includegraphics[width=0.98\textwidth]{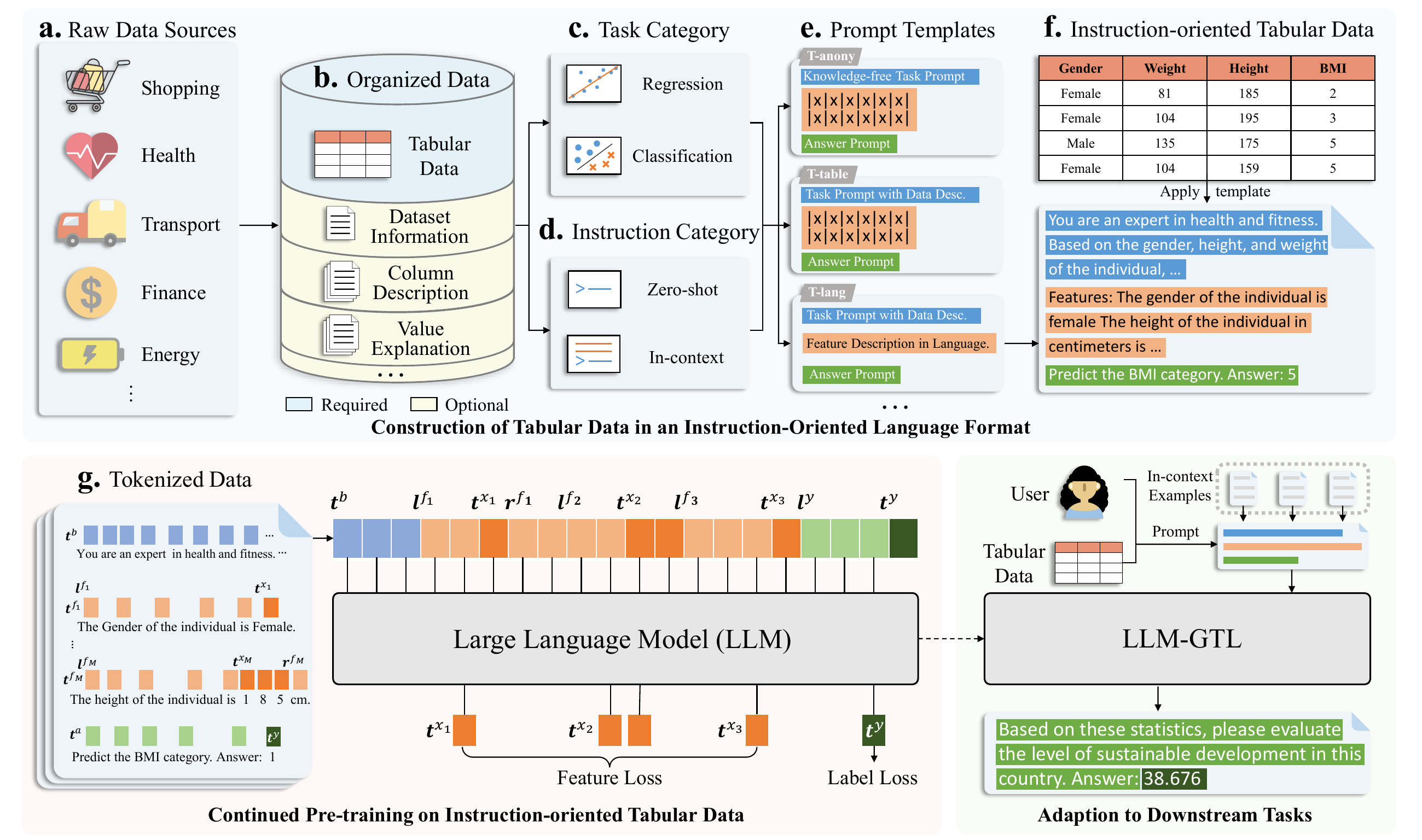}
    \caption{An overview of generative tabuar learning for LLMs.}
    \label{fig:framework}
\end{figure*}

\section{Generative Tabular Learning for Large Language Models}
\label{sec:method}

Figure~\ref{fig:framework} provides a comprehensive overview of our generative paradigm for tabular deep learning with LLMs, including the pipeline for constructing tabular data in an instruction-oriented language format across various domains (Section~\ref{sec:method_data}), the detailed optimization process in GTL (Section~\ref{sec:method_GTL}), and an example of the prompt-based adaptation of the GTL-guided LLM to downstream tasks.

\subsection{Problem Formulation}
\label{sec:method_prob}

\paragraph{Notations}
In the field of tabular learning, we typically handle a tabular task $\mathcal{T}: \mathcal{X} \rightarrow \mathcal{Y}$, which associates a tabular instance $\bm{x} \in \mathcal{X}$, consisting of $M$ features, $\{x_i\}_{i=1}^M$, with a prediction target, $y \in \mathcal{Y}$. For regression tasks, $\mathcal{Y} \subseteq \R^1$ and for $C$-class classification tasks, $\mathcal{Y} = \{0, 1, \cdots, C-1\}$. Note that tabular features are generally divided into numerical and categorical types, while we do not explicitly distinguish between them here. Besides, each tabular task may also be associated with various meta-information elements, such as the task background, prediction target interpretation, and feature descriptions. We denote the comprehensive meta-information of task $\mathcal{T}$ as $\mathcal{M}^{\mathcal{T}}$.

Traditional tabular data learning methods primarily focus on constructing a discriminative model to learn the dependency, $p(y|\bm{x})$, between the target and features using the training data. However, such a model strongly tied to the training data faces significant challenges when adapting to a new tabular task, $\tilde{\mathcal{T}}$, with a different data schema $\tilde{\mathcal{X}}$ and prediction targets $\tilde{\mathcal{Y}}$.

\paragraph{Zero-shot Learning}
In the realm of tabular data, zero-shot learning~\citep{hegselmann2023TabLLM,yang2023UniTabE} is typically characterized by the ability to predict outcomes for data samples from a previously unseen task, $\tilde{\mathcal{T}}$. This new task possesses an unknown data schema, $\tilde{\mathcal{X}}$, but it has meta-information $\mathcal{M}^{\tilde{\mathcal{T}}}$, which encapsulates the meaning of features and prediction targets. Formally, we define the input of a zero-shot learning task as a tuple $(\mathcal{M}^{\tilde{\mathcal{T}}}, \tilde{\bm{x}}^{new})$, where $\mathcal{M}^{\tilde{\mathcal{T}}}$ denotes the meta-information for the new task and $\tilde{\bm{x}}^{new}$ represents a novel data sample for which we aim to make predictions.

\paragraph{In-context Learning (Few-shot Learning)}
In-context learning resembles zero-shot learning but with an additional dimension: it includes a set of exemplars from the new task $\tilde{\mathcal{T}}$ as demonstrations, represented as $D^{\tilde{\mathcal{T}}} = \{ (\tilde{\bm{x}}^j, \tilde{y}^j)\}_{j=1}^N$, where $N$ signifies the quantity of in-context examples. We formally characterize the input of an in-context learning task as a tuple $(\mathcal{M}^{\tilde{\mathcal{T}}}, D^{\tilde{\mathcal{T}}}, \tilde{\bm{x}}^{new})$, where the meta-information $\mathcal{M}^{\tilde{\mathcal{T}}}$ can be omitted if the model does not utilize it, as in the case of TabPFN~\citep{hollmann2022TabPFN}. In-context learning can seamlessly transition into few-shot learning by allowing the model to undergo further fine-tuning on $D^{\tilde{\mathcal{T}}}$ before making predictions on $\tilde{\bm{x}}^{new}$.

\subsection{Construction of Tabular Data in an Instruction-Oriented Language Format}
\label{sec:method_data}

The upper part of Figure~\ref{fig:framework} depicts the pipeline for our data construction.
Our preliminary step is to collect a significant amount of tabular datasets, which cover a broad spectrum of predictive tasks across multiple domains. This process, with more details in Section~\ref{sec:exp_setup}, requires the acquisition of both features and labels, better including meta-information (optional).
Then, for each specified tabular task, $\mathcal{T}$, and any accompanying meta information $\mathcal{M}^{\mathcal{T}}$, it is essential to transform the data samples from their original tabular format into an instruction-oriented language format to support the subsequent GTL procedure.

This instruction-oriented format primarily consists of three components. The first component pertains to task instructions, detailing the background and the objective of this prediction task. The second component offers detailed descriptions of the features, encompassing both their values and associated meanings (when available). It's important to highlight that we have the ability to incorporate in-context examples within this component, preceding the current data sample. Additionally, we can introduce a prompt indicating the presence of in-context demonstrations for reference. The third component encompasses answer descriptions, typically prefaced with an answer instruction, followed by the prediction targets.

Adhering to these guidelines, we devise various templates that cater to different ways of expressing feature descriptions and varying requirements.

Initially, we formulate the \texttt{T-lang} template, which delineates tabular features in a manner akin to natural language, similar to previous research in feeding tabular data to LLMs~\citep{hegselmann2023TabLLM,dinh2022LIFT}. The major advantage of this template is that it emulates human language style, facilitating smoother knowledge transfer for LLMs, given their extensive training on human language data. A language-style description of a tabular data instance may induce improved performance. Detailed composition of this template is provided in Appendix~\ref{app:temp_lang}. 
However, it is important to note that this template falls short in efficiently integrating in-context examples, as it necessitates the conversion of each example into a description, resulting in repetitive descriptions for feature meanings. 

To address this limitation, we introduce the \texttt{T-table} template, which employs a Markdown table format to encapsulate tabular feature values. This approach ensures that irrespective of the number of in-context examples added, we only need to specify a single table header containing brief tags for different features. Feature meanings are positioned ahead of this Markdown table. The \texttt{T-table} template efficiently handles tabular data, particularly in the in-context learning setup. However, it also poses challenges for the long-term association capabilities of LLMs, as they inherently need to link a feature value with its corresponding table column and subsequently associate it with the correct feature meaning explanations. Detailed composition of this template is provided in Appendix~\ref{app:temp_table}.

The \texttt{T-anony} template, a derivative of the \texttt{T-table} template, omits all meta-information. This is designed to simulate a practical scenario where we possess tabular data samples, but lack knowledge of their background or feature meanings. This setup is more akin to TabPFN~\citep{hollmann2022TabPFN}, which did not utilize semantic information about feature columns and task background. Detailed composition of this template is provided in Appendix~\ref{app:temp_anony}.

By employing these templates, we can adapt to an array of tabular tasks and data schemas across diverse domains. Additionally, we have the capability to utilize meta information associated with these tabular datasets when available, and also have mechanisms in place for situations where such information is absent. Furthermore, we can regulate the number of in-context examples incorporated, thereby effortlessly incorporating both zero-shot and in-context learning scenarios.

\subsection{Learning and Adaptation}
\label{sec:method_GTL}

The bottom part of Figure~\ref{fig:framework} intuitively depicts the training and adaptation of GTL. Next, we formally introduce the GTL paradigm.

\paragraph{Notations for Tokenized Instruction-oriented Language Data}
Below we mainly leverage the \texttt{T-lang} template to derive the notations for a tokenized instruction-oriented language sample.
Recalling in the last section, an instruction-oriented language format for tabular data primarily incorporates three components, encompassing a task instruction, a feature description, and an answer prompt.
Formally, we represent the tokenization results of the first part as: $\bm{t}^b = [t^b_1, \cdots, t^b_{|\bm{t}^b|}]$, where $|\cdot|$ is the operation to denote the length of a token sequence.
Besides, we denote the $i$-th feature as $\bm{t}^{f_i} = [\bm{l}^{f_i}, \bm{t}^{x_i}, \bm{r}^{f_i}]$,
where $\bm{l}^{f_i} = [l^{f_i}_1, \cdots, l^{f_i}_{|\bm{l}^{f_i}|}]$ includes all tokens of feature descriptions on the left of this feature value,
$\bm{t}^{x_i} = [t^{x_i}_1, \cdots, t^{x_i}_{|\bm{t}^{x_i}|}]$ denotes the token sequence for the feature value $x_i$,
and $\bm{r}^{f_i} = [r^{f_i}_1, \cdots, r^{f_i}_{|\bm{r}^{f_i}|}]$ includes the tokens for remaining feature descriptions on the right of feature values.
By concatenating token sequences for all features, we have the overall feature sequence as $\bm{t}^f = [\bm{t}^{f_1}, \cdots, \bm{t}^{f_M}]$.
Last, we represent the tokenized answer prompt as $\bm{t}^a = [\bm{l}^a, \bm{t}^{y}]$,
where $\bm{l}^a = [l^a_1, \cdots, l^a_{|\bm{l}^a|}]$ denotes the answer prompt before the answer tokens $\bm{t}^{y} = [t^y_1, \cdots, t^y_{|\bm{t}^y|}]$.
In this way, we can express a tabular instance $\bm{x}$ and the associated target $y$ as a sequence of tokens:
\begin{align}
    [\bm{t}^b, \bm{t}^f, \bm{t}^a] = [\bm{t}^b, \bm{l}^{f_1}, \bm{t}^{x_1}, \bm{r}^{f_1}, \cdots, \bm{l}^{f_M}, \bm{t}^{x_M}, \bm{r}^{f_M}, \bm{l}^a, \bm{t}^y],
    \label{eq:tab_text_repre}
\end{align}
which systematically unifies task background ($\bm{t}^b$), feature meanings ($\{\bm{l}^{f_i}, \bm{r}^{f_i}\}_{i=1}^M$), and feature values ($\{\bm{t}^{x_i}\}_{i=1}^M$) and supports various prediction targets via a variable-length sequence ($\bm{t}^y$).
Please note that, the above tokenization notations also apply for other templates, such as \texttt{T-table}, merely with a different way of specifying feature meanings.
Besides, we need to mark the positions of feature value tokens and prediction target tokens of the current tabular data sample to support the subsequent GTL procedure.

\paragraph{The Objective of GTL}
Based on the notations for tokenized data, we devise GTL as characterizing the following joint distribution:
\begin{align}
p(\bm{x}, y) = p(\bm{t}^{\bm{x}}, \bm{t}^y | \bm{t}^m) = p(\bm{t}^y | \bm{t}^{\bm{x}}, \bm{t}^{m}) \prod_{i=1}^M p (\bm{t}^{x_i} | \bm{t}^{< x_i}),
\label{eq:decouple_joint_prob}
\end{align}
where we introduce additional notations to ensure the concise formulation, using $\bm{t}^{m} = [\bm{t}^{b}, \bm{l}^{f_1}, \bm{r}^{f_1}, \cdots, \bm{l}^{f_M}, \bm{r}^{f_M}, \bm{l}^{a}]$ to denote all meta information, $\bm{t}^{\bm{x}} = [\bm{t}^{x_1}, \cdots, \bm{t}^{x_M}]$ to represent all tokens related to feature values, and $\bm{t}^{< x_i}$ to include all tokens ahead of $\bm{t}^{x_i}$ in~\eqref{eq:tab_text_repre}.
Here $p(\bm{x}, y)$ denotes the joint distribution in the initial feature and label spaces, while $p(\bm{t}^{\bm{x}}, \bm{t}^y | \bm{t}^{m})$ represents the same joint distribution conditioned on all meta information using the text representation, which can further be decoupled autoregressively into $p(\bm{t}^y | \bm{t}^{\bm{x}}, \bm{t}^{m}) \prod_{i=1}^M p (\bm{t}^{x_i} | \bm{t}^{< x_i})$.
It is thus straightforward to leverage LLMs, especially those using auto-regressive architectures, to characterize this decoupled formulation.
The only modification needed is to mask out the losses on meta-information tokens.
While GTL employs the next-token prediction loss akin to LLMs, it distinguishes itself from auto-regressive pre-training on language data. Specifically, GTL explicitly stimulates LLMs to discern complex dependencies between prediction target tokens and feature tokens. It encourages the capture of intricate relationships between current features and in-context examples, and fosters the establishment of effective connections between diverse language instructions and numerical data.

\paragraph{Adaptation to Downstream Tasks.}
Given an LLM, we refer to its variant after the GTL process as LLM-GTL. When adapting to a new task, irrespective of the data schema or task type, LLM-GTL facilitates direct inference by simply specifying a prompt.
For instance, if the objective is to optimally utilize prior knowledge for zero-shot inference on semantically-rich tasks, the \texttt{T-lang} template can be used to transform data samples, which are then fed into the LLM-GTL for output generation.
In situations that necessitate the inclusion of more in-context examples and also require leveraging meta information about the tabular task, the \texttt{T-table} template emerges as the best practice.
Furthermore, even in the absence of meta information, LLM-GTL can still offer predictions via the \texttt{T-anony} template, relying on the inherent statistical learning acquired during the GTL process.


%% file: KDD2024_Camera/4_experiments.tex
\section{Experiments}
\label{sec:exp}

This section presents extensive empirical investigations to address the following key research questions:
1) How does the application of GTL to an LLM impact its convergence and generalization behaviors on tabular deep learning? 2) What are the crucial scaling laws associated with GTL? 3) How do different text templates influence the performance? 4) How do GTL-enhanced LLMs compare against both traditional tabular models and contemporary LLMs? 

\subsection{Experimental Setups}
\label{sec:exp_setup}

\paragraph{Dataset Collection}
We have curated a collection of 384 public tabular datasets from Kaggle\footnote{\url{https://www.kaggle.com/datasets}}, which includes 176 classification and 208 regression tasks spanning a wide range of industrial domains. For the purpose of holdout evaluation, we have randomly selected 44 tasks, consisting of 20 classification and 24 regression tasks, and carefully examined to ensure no overlap with the remaining tasks utilized for the continued pre-training with GTL. Further details pertaining to the statistics of the datasets and the covered domains can be found in Appendix~\ref{app:data_domain_dist}.

\paragraph{Configurations for Pre-training}
For the 340 tabular datasets allocated for continued pre-training, we examine six different scenarios of zero-shot or in-context learning, with the number of in-context examples in $\{0, 4, 8, 16, 32, 64\}$. 
To further enrich the diversity of learning experiences, we have created up to four unique tasks for each dataset by judiciously selecting various columns as labels and the remainder as features. 
Each scenario is tested with all three text templates, resulting in a total of 14k cases. Typically, we randomly select $64$ data samples for each case, obtaining 880k tokenized sequences for GTL. We impose a maximum sequence length of $4,096$ and discard any data samples that exceed this limit, yielding a total of 640k data samples for pre-training. We also sample a small subset of test samples, distinct from these pre-training samples, to assess in-domain generalization on these pre-training datasets across different instruction categories. We sample $16$ per classification case and $4$ per regression case (the latter requiring multi-step decoding which is highly time-consuming during evaluation).
We have utilized the 7B and 13B versions of LLaMA 2~\citep{touvron2023llama2} as our base LLMs. The corresponding GTL-enhanced variants are denoted as LLaMA-7B-GTL and LLaMA-13B-GTL, respectively. Implementation details can be found in Appendix~\ref{app:imp_details_GTL}.

\paragraph{Configurations for Holdout Evaluation}
For the 44 holdout datasets, we adopt a similar approach to that of pre-training data, considering different in-context examples and text templates. To achieve a robust evaluation that mitigates the effects of randomness, we use three distinct random seeds to sample evaluation examples for each case, where one case is a combination of a dataset, a in-context configuration, and a text template. We increase the sample size to $64$ for classification cases and $16$ for regression cases. By observing performance variations across different seeds within the same case, we ascertain that this configuration balances the computational cost of running diverse experiments with the statistical significance of the results obtained. In total, we have around 88k data samples for holdout evaluation. The primary metrics used for evaluation are the Area Under the Receiver Operating Characteristic (AUROC) for classification tasks and the Normalized Mean Absolute Error (NMAE) for regression tasks.

\paragraph{Baselines.}
Our baseline includes both competitive tabular models and contemporary large language models (LLMs). For the tabular models, we have included classic tree-based models such as XGBoost~\citep{chen2016XGBoost}, CatBoost~\citep{prokhorenkova2018catboost}, and LightGBM~\citep{ke2017LightGBM}. We have also considered the state-of-the-art in-context model, TabPFN~\citep{hollmann2022TabPFN}, a competitive neural model, FTT~\citep{gorishniy2021revisit_tab_dnn}, and a logistic regression (LR) baseline, known for its robustness in extremely few-shot scenarios. As for the LLMs, we have included the initial versions of LLaMA 2, denoted as LLaMA-7B and LLaMA-13B, and proprietary LLMs such as GPT-3.5~\citep{Brown2020GPT-3} and GPT-4~\citep{OpenAI2023GPT4TR}. We assessed the proprietary models by invoking their APIs on our holdout data. To calculate AUROC using prediction probabilities, we instructed these models to output probabilities. For LLaMA and GTL models, due to our access to their networks, we instructed them to predict class indices directly and collected logits over the class tokens to acquire output probabilities. For the NMAE metric, all LLMs require multi-step decoding, for which we allowed a maximum of ten steps for LLaMA models, sufficient for all cases under consideration in this study.


\subsection{Understanding GTL}
\label{sec:exp_understand_GTL}


\begin{figure*}[tb]
    \centering
    \includegraphics[width=0.98\textwidth]{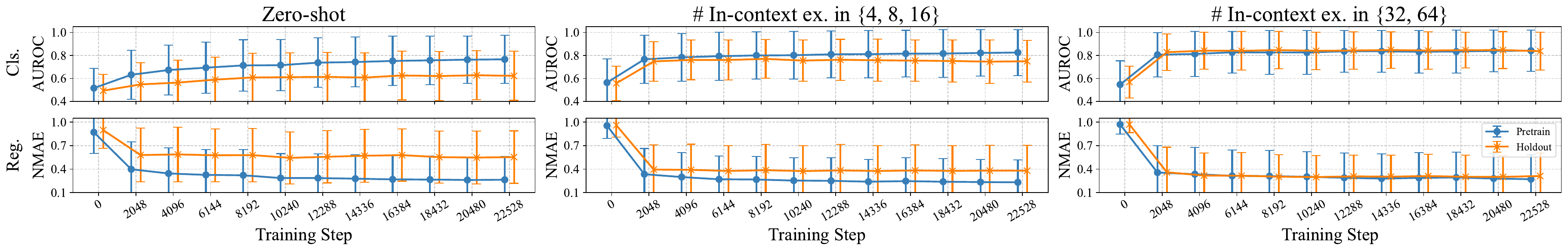}
    \caption{An examination of in-domain (on pre-train test sets) and out-of-domain (on holdout data) generalization as GTL optimization progresses, highlighting the zero-shot learning scenario (left), the in-context learning scenario with few in-context examples (middle), and with many examples (right).}
    \label{fig:convergence}
\end{figure*}

\begin{figure*}[tb]
    \centering
    \includegraphics[width=0.98\textwidth]{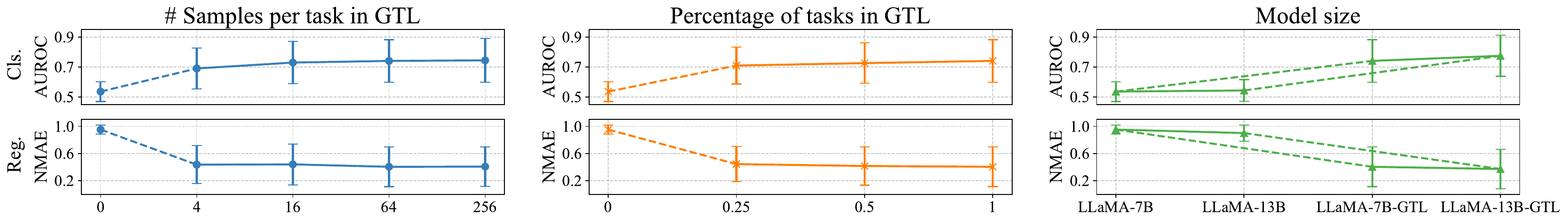}
    \caption{A study of scaling laws, focusing on the number of data samples per pre-training task (left), the percentage of pre-training tasks (middle), and the model size (right). Both AUROC and NMAE metrics, aggregated from all holdout datasets across different context examples, are represented. A dotted line connects a LLaMA model with its GTL variant for comparison.}
    \label{fig:scaling_law}
\end{figure*}

\begin{figure}[tb]
    \includegraphics[width=0.98\columnwidth]{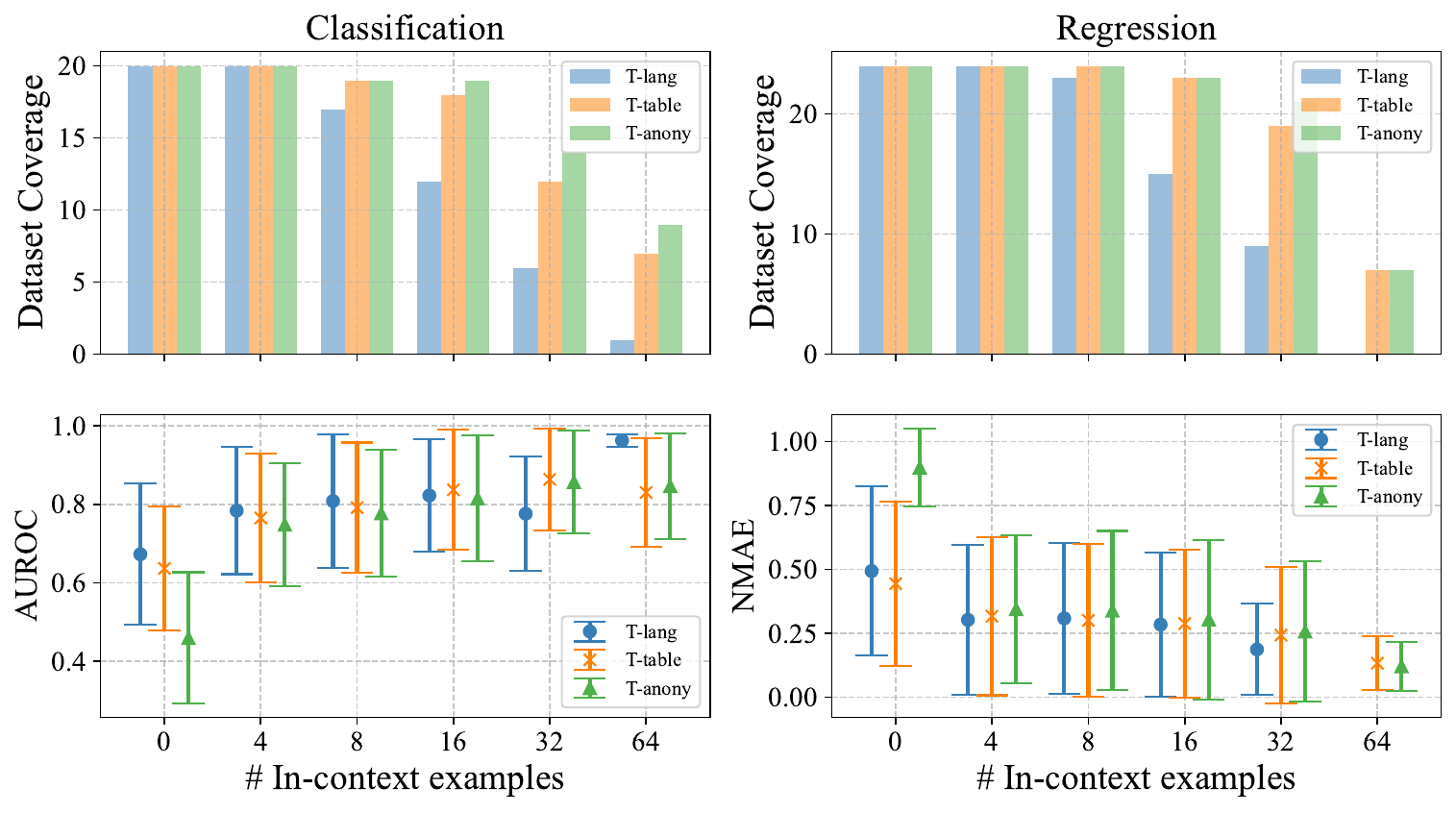}
    \caption{The influence of template variability on dataset coverage, given a maximum sequence length, and on generalization performance, considering different numbers of in-context examples on holdout datasets.}
    \label{fig:template}
\end{figure}

We investigate the generalization behaviors during the optimization of GTL, as depicted in Figure~\ref{fig:convergence}. Notably, a significant gap emerges between the improvements of in-domain and out-of-domain generalization in the zero-shot scenario. Initially, both generalization abilities enhance swiftly, but as training progresses, in-domain generalization continues to improve while out-of-domain generalization plateaus. This plateau occurs due to the presence of domain-specific knowledge in holdout data, which is not covered by either the LLM's prior understanding or pre-train data. Consequently, early training steps primarily improve tabular data understanding. As training deepens, the model continues to learn patterns within pre-train domains but acquires no further universal abilities for holdout datasets. However, the introduction of in-context examples narrows this gap. Specifically, when more than 32 in-context examples are included, the gap between in-domain and out-of-domain generalization nearly disappears, whereas they roughly converge after $8,192$ steps. This analysis suggests that early stopping of GTL optimization can be beneficial for generalization on unseen datasets and domains. Therefore, in subsequent experiments on scaling laws, we halt pre-training at step $8,192$.



The study of scaling laws seeks to identify key factors influencing generalization to holdout data, which could indicate potential avenues for future breakthroughs. We primarily focus on the impact of scaling the number of data samples per task, the number of tasks used for GTL, and the model size. As shown in Figure~\ref{fig:scaling_law}, scaling the number of data samples per tabular task improves generalization quickly when LLMs have limited exposure to tabular data. However, beyond a certain threshold, such as 64, further increases to 256 do not significantly enhance holdout data generalization. In contrast, scaling the number of tasks consistently improves generalization, suggesting that expanding task coverage beyond the 340 datasets used for pre-training here could further enhance GTL performance. Moreover, increasing the model size, for example from 7B to 13B, significantly boosts performance. 


Additionally, we thoroughly compare the effects of different data templates on holdout generalization across various in-context scenarios. As Figure~\ref{fig:template} depicts, the \texttt{T-lang} template, which best adheres to language conventions, excels in semantic-dominant scenarios, particularly zero-shot and few-shot classification tasks. However, due to its inefficient token usage, it becomes less effective as the number of in-context examples increase, with a significant drop-off observed at 16 examples. In contrast, regression tasks, which primarily involve fitting a numerical target based on different features, do not heavily rely on semantic information. Based on these findings, we propose a template usage strategy: \texttt{T-lang} for zero-shot classification cases, \texttt{T-table} for other scenarios when meta information is available, and \texttt{T-anony} when the origin of the tabular data is unknown.

\subsection{Performance Comparisons with Competitive Tabular Models and LLMs}
\label{sec:exp_compare}

\input{KDD2024_Camera/tables/main_results}
We further contrast our approach with competitive tabular models and contemporary LLMs, as illustrated in Table~\ref{tab:cls_results} and Table~\ref{tab:reg_results}.  


\emph{When compared to tabular models, our approach excels in numerous few-shot cases, including classification tasks with in-context examples ranging from 4 to 16, and regression tasks with examples from 4 to 32.}
These findings validate the effectiveness of the GTL paradigm in integrating prior knowledge and tabular data understanding, a fusion that could be invaluable in addressing data-scarce challenges in many practical applications, such as rare disease classification or new material property prediction. A closer look at the classification results reveals TabPFN to be a robust baseline, outperforming us in average rank due to its superior handling of larger in-context examples. 
Regarding regression tasks, our untuned LLaMA-13B-GTL model frequently surpasses other models that have been tuned on in-context examples, also demonstrating the effectiveness of GTL. Besides, we note a rise in error with our model when 64 contexts are used, potentially due to the limited inclusion of such examples in pre-training regression tasks. We expect this limitation can be mitigated by scaling up the maximum sequence length, model size, and task quantity. 

\emph{Our GTL-enhanced LLMs, potentially utilizing smaller model sizes, surpass GPT-4 in numerous classification scenarios and significantly narrow the performance gap in regression tasks.}
GPT-4, being proprietary with undisclosed pre-training data, raises potential concerns of data contamination since all tabular datasets used in our study are publicly available on Kaggle. Despite this potential issue, the exceptional performance exhibited by our LLaMA-GTL series, derived from open-source LLMs, affirms the effectiveness of the GTL paradigm in bolstering tabular deep learning capabilities for LLMs. Beyond the realm of tabular data, our findings suggest that open-source LLMs should integrate instruction-oriented tabular data into their pre-training corpus. This could amplify their data comprehension capabilities, especially considering the ubiquity of tabular data across various fields, and possibly spark new discoveries through the amalgamation of diverse data types. Finally, we conjecture that the superior performance of our GTL methodology over GPT-4 in classification tasks might be ascribed to different approaches in prediction probability computation, as we are limited to accessing GPT-4's prediction probabilities via instruction-based API calls.





%% file: KDD2024_Camera/tables/main_results.tex
\begin{table*}[t]
\centering
\scriptsize
\caption{Classification results. We display the mean, 25th percentile, and 75th percentile AUROC scores across all datasets for both zero-shot and in-context learning scenarios. $\text{Rank}_{z}$  denotes the average rank in zero-shot settings, while $\text{Rank}_{i}$ signifies the average rank in in-context settings.}
\label{tab:cls_results}
\resizebox{\textwidth}{!}{
\begin{tabular}{ccccccccc}
\toprule
& 0 (Zero-shot) & 4 & 8 & 16 & 32 & 64 & $\text{Rank}_{z}$ & $\text{Rank}_{i}$ \\
\midrule
LightGBM & $-$ & $0.500_{[0.500, 0.500]}$ & $0.500_{[0.500, 0.500]}$ & $0.500_{[0.500, 0.500]}$ & $0.500_{[0.500, 0.500]}$ & $\underline{0.848_{[0.716, 0.965]}}$ & $-$ & $9.754$ \\
XGBoost & $-$ & $0.555_{[0.500, 0.543]}$ & $0.685_{[0.500, 0.799]}$ & $0.771_{[0.671, 0.941]}$ & $0.834_{[0.710, 0.959]}$ & $0.834_{[0.767, 0.963]}$ & $-$ & $6.174$ \\
CatBoost & $-$ & $\underline{0.727_{[0.612, 0.882]}}$ & $0.774_{[0.675, 0.901]}$ & $0.802_{[0.735, 0.930]}$ & $\bm{0.853_{[0.746, 0.973]}}$ & $0.844_{[0.778, 0.956]}$ & $-$ & $4.272$ \\
\midrule
FTTransformer & $-$ & $0.669_{[0.480, 0.815]}$ & $0.673_{[0.542, 0.844]}$ & $0.679_{[0.533, 0.785]}$ & $0.709_{[0.594, 0.820]}$ & $0.699_{[0.609, 0.773]}$ & $-$ & $7.564$ \\
XTab & $-$ & $0.588_{[0.457, 0.734]}$ & $0.597_{[0.505, 0.678]}$ & $0.671_{[0.563, 0.769]}$ & $0.759_{[0.649, 0.884]}$ & $0.767_{[0.666, 0.941]}$ & $-$ & $7.554$ \\
LR & $-$ & $0.715_{[0.542, 0.886]}$ & $0.743_{[0.630, 0.889]}$ & $0.800_{[0.725, 0.925]}$ & $0.850_{[0.759, 0.965]}$ & $0.813_{[0.700, 0.954]}$ & $-$ & $4.308$ \\
TabPFN & $-$ & $0.722_{[0.604, 0.884]}$ & $0.758_{[0.634, 0.872]}$ & $\underline{0.803_{[0.733, 0.933]}}$ & $\underline{0.852_{[0.747, 0.970]}}$ & $\bm{0.850_{[0.787, 0.946]}}$ & $-$ & $\underline{4.236}$ \\
\midrule
LLaMA-7B & $0.504_{[0.457, 0.585]}$ & $0.545_{[0.451, 0.622]}$ & $0.525_{[0.457, 0.585]}$ & $0.494_{[0.429, 0.556]}$ & $0.493_{[0.430, 0.545]}$ & $0.500_{[0.461, 0.557]}$ & $4.000$ & $10.251$ \\
GPT-3.5 & $0.539_{[0.456, 0.632]}$ & $0.525_{[0.490, 0.576]}$ & $0.484_{[0.466, 0.541]}$ & $0.536_{[0.485, 0.571]}$ & $0.509_{[0.455, 0.552]}$ & $0.520_{[0.483, 0.549]}$ & $3.900$ & $10.097$ \\
LLaMA-13B & $0.495_{[0.450, 0.566]}$ & $0.507_{[0.436, 0.573]}$ & $0.555_{[0.461, 0.582]}$ & $0.520_{[0.461, 0.590]}$ & $0.487_{[0.446, 0.555]}$ & $0.521_{[0.472, 0.562]}$ & $4.550$ & $9.990$ \\
GPT-4 & $\bm{0.654_{[0.537, 0.824]}}$ & $0.642_{[0.563, 0.785]}$ & $0.721_{[0.596, 0.887]}$ & $0.713_{[0.573, 0.832]}$ & $0.711_{[0.570, 0.866]}$ & $0.658_{[0.531, 0.753]}$ & $\bm{2.450}$ & $6.805$ \\
\midrule
LLaMA-7B-GTL & $0.579_{[0.473, 0.679]}$ & $0.714_{[0.581, 0.847]}$ & $\underline{0.777_{[0.649, 0.927]}}$ & $0.793_{[0.656, 0.940]}$ & $0.821_{[0.728, 0.936]}$ & $0.813_{[0.720, 0.933]}$ & $3.550$ & $5.164$ \\
LLaMA-13B-GTL & $\underline{0.641_{[0.565, 0.752]}}$ & $\bm{0.767_{[0.663, 0.919]}}$ & $\bm{0.811_{[0.681, 0.960]}}$ & $\bm{0.823_{[0.752, 0.955]}}$ & $0.850_{[0.777, 0.955]}$ & $0.816_{[0.733, 0.918]}$ & $\underline{2.500}$ & $\bm{3.903}$ \\
\bottomrule
\end{tabular}}
\end{table*}

\begin{table*}[t]
\centering
\scriptsize
\caption{Regression results. We display the mean, 25th percentile, and 75th percentile NMAE (Normalized Mean Absolute Error) scores across all datasets for both zero-shot and in-context learning scenarios. NMAE values are truncated at a maximum of 1.0 to account for certain baselines' diminished performance due to inadequate training data. $\text{Rank}_{z}$ denotes the average rank in zero-shot settings, while $\text{Rank}_{i}$ signifies the average rank in in-context settings.}
\label{tab:reg_results}
\resizebox{\textwidth}{!}{
\begin{tabular}{ccccccccc}
\toprule
& 0 (Zero-shot) & 4 & 8 & 16 & 32 & 64 & $\text{Rank}_{z}$ & $\text{Rank}_{i}$ \\
\midrule
LightGBM & $-$ & $0.422_{[0.220, 0.534]}$ & $0.414_{[0.199, 0.540]}$ & $0.436_{[0.226, 0.554]}$ & $0.439_{[0.191, 0.550]}$ & $0.281_{[0.091, 0.252]}$ & $-$ & $7.304$ \\
XGBoost & $-$ & $0.376_{[0.154, 0.475]}$ & $0.350_{[0.134, 0.512]}$ & $0.348_{[0.116, 0.507]}$ & $0.284_{[0.054, 0.400]}$ & $\bm{0.141_{[0.029, 0.161]}}$ & $-$ & $5.507$ \\
CatBoost & $-$ & $0.372_{[0.165, 0.520]}$ & $0.341_{[0.142, 0.459]}$ & $0.337_{[0.115, 0.434]}$ & $0.313_{[0.069, 0.405]}$ & $0.224_{[0.040, 0.184]}$ & $-$ & $4.996$ \\
\midrule
XTab & $-$ & $0.504_{[0.251, 0.779]}$ & $0.419_{[0.166, 0.634]}$ & $0.398_{[0.157, 0.567]}$ & $0.400_{[0.132, 0.732]}$ & $0.275_{[0.097, 0.263]}$ & $-$ & $7.386$ \\
FTTransformer & $-$ & $0.414_{[0.206, 0.514]}$ & $0.407_{[0.201, 0.509]}$ & $0.433_{[0.221, 0.572]}$ & $0.424_{[0.189, 0.610]}$ & $0.355_{[0.180, 0.362]}$ & $-$ & $7.346$ \\
LR & $-$ & $0.384_{[0.151, 0.540]}$ & $0.373_{[0.104, 0.595]}$ & $0.358_{[0.058, 0.553]}$ & $0.382_{[0.041, 0.927]}$ & $0.211_{[0.018, 0.164]}$ & $-$ & $5.621$ \\
\midrule
LLaMA-7B & $0.938_{[0.977, 1.000]}$ & $0.902_{[0.945, 1.000]}$ & $0.954_{[1.000, 1.000]}$ & $0.995_{[1.000, 1.000]}$ & $0.987_{[1.000, 1.000]}$ & $0.956_{[0.988, 1.000]}$ & $5.250$ & $10.604$ \\
LLaMA-13B & $0.770_{[0.603, 1.000]}$ & $0.792_{[0.535, 1.000]}$ & $0.844_{[0.826, 1.000]}$ & $0.873_{[0.999, 1.000]}$ & $0.938_{[0.999, 1.000]}$ & $0.774_{[0.429, 1.000]}$ & $4.167$ & $9.932$ \\
GPT-3.5 & $\underline{0.492_{[0.199, 0.842]}}$ & $0.393_{[0.183, 0.470]}$ & $0.406_{[0.217, 0.503]}$ & $0.426_{[0.211, 0.530]}$ & $0.408_{[0.179, 0.506]}$ & $0.325_{[0.157, 0.410]}$ & $3.083$ & $7.114$ \\
GPT-4 & $0.583_{[0.241, 0.958]}$ & $\bm{0.229_{[0.083, 0.304]}}$ & $\bm{0.233_{[0.072, 0.297]}}$ & $\bm{0.229_{[0.064, 0.286]}}$ & $\bm{0.214_{[0.054, 0.268]}}$ & $\underline{0.146_{[0.030, 0.240]}}$ & $3.250$ & $\bm{2.918}$ \\
\midrule
LLaMA-7B-GTL & $0.523_{[0.223, 0.957]}$ & $0.316_{[0.102, 0.417]}$ & $0.299_{[0.082, 0.387]}$ & $0.311_{[0.081, 0.513]}$ & $0.291_{[0.068, 0.511]}$ & $0.229_{[0.031, 0.315]}$ & $\underline{2.792}$ & $4.589$ \\
LLaMA-13B-GTL & $\bm{0.430_{[0.172, 0.671]}}$ & $\underline{0.288_{[0.079, 0.403]}}$ & $\underline{0.271_{[0.059, 0.398]}}$ & $\underline{0.295_{[0.063, 0.415]}}$ & $\underline{0.260_{[0.050, 0.355]}}$ & $0.165_{[0.035, 0.263]}$ & $\bm{2.083}$ & $\underline{3.946}$ \\
\bottomrule
\end{tabular}}
\end{table*}

%% file: KDD2024_Camera/5_conclusion.tex
\section{Conclusion}
\label{sec:conclusion}

This study introduces GTL, an generative paradigm that fuses LLM's instruction-following capabilities with tabular deep learning. The preliminary results demonstrate its effectiveness, particularly in tackling data scarcity issues, and our scaling analyses suggest avenues for further performance enhancements.

\paragraph{Limitations}
Despite these encouraging outcomes, our study does have limitations. The context length constraint of LLMs restricts the number of in-context examples we can use, which can hinder learning over large-scale data. Besides, using a text format to represent tabular data is not token-usage efficient. Furthermore, the computational cost of LLMs is considerably higher than traditional techniques, raising financial and environmental concerns. Lastly, despite using 384 datasets in this study, the diversity and scale may not fully represent the potential and applicability of GTL across all tabular data fields.

\paragraph{Future Directions}
In the meanwhile, we are optimistic about this research direction's future prospects. For example, the GTL paradigm opens doors for conversational tabular deep learning, where models could refine predictions through dialogue. GTL could also promote interpretable learning over tabular data, as LLMs can generate explanations with predictions, albeit with challenges in ensuring explanation faithfulness. There is potential for better integration between LLMs and traditional models, merging their strengths. Furthermore, our work provides a valuable asset for the LLM community, serving both as a benchmark for evaluating data comprehension capabilities and as a source to enrich existing pre-training corpora.

%% file: KDD2024_Camera/6_appendix.tex

\section{Data}
\label{app:data_appen}


\subsection{Detailed Statistics of Collected Datasets}
\label{app:data_stats}

To empower Large Language Models (LLMs) to build foundational knowledge and abilities across a variety of domains through generative tabular learning, the compilation of a diverse and substantial collection of tabular datasets is pivotal. We have carefully curated a collection of 384 public datasets, comprised of high-quality, functional tabular classification datasets procured from Kaggle. 

\subsection{Domain Distribution}
\label{app:data_domain_dist}
\begin{figure}[b]
    \centering
    \includegraphics[width=0.9\columnwidth]{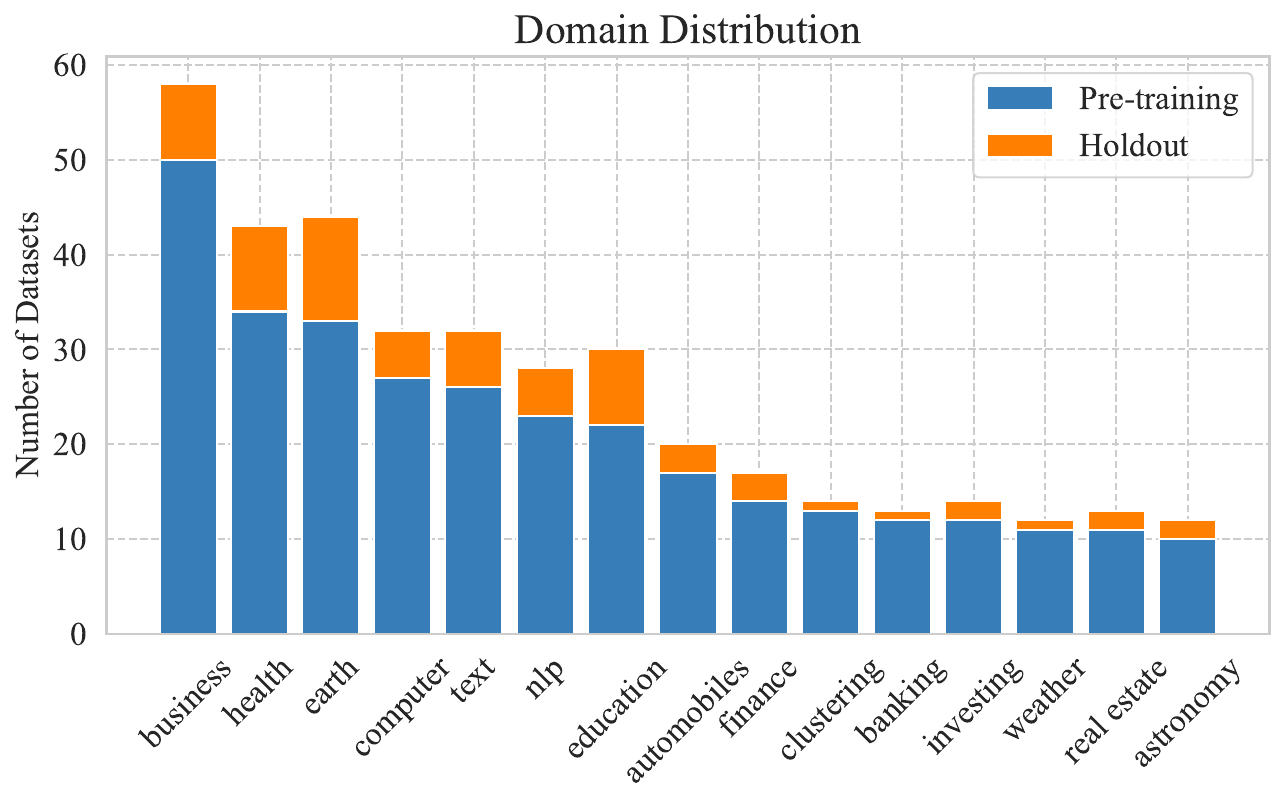}
    \caption{Domain distribution of pre-training and holdout datasets.}
    \label{fig:domain_dist}
\end{figure}

We have randomly divided all the acquired Kaggle datasets into pre-training and holdout sets, ensuring the preservation of domain distribution. The top 15 domain tags, along with the associated number of datasets for each domain, are displayed in Figure~\ref{fig:domain_dist}.

\subsection{Template}
\label{app:temp_design}

\subsubsection{The \texttt{T-lang} Template}
\label{app:temp_lang}
The T-lang template preserves the semantic information of data samples by converting each feature into a sentence. 

\subsubsection{The \texttt{T-table} Template}
\label{app:temp_table}
The T-table template maintains the tabular format of the data samples. Meta information, such as feature description and label description, is added prior to the data and is not repeated with each example when in-context examples are introduced. 
An instance of the T-table template with in-context examples is showcased in List \ref{lst:t_table_cont_eg}.

\subsubsection{The \texttt{T-anony} Template}
\label{app:temp_anony}
The T-anony template eliminates all meta information pertaining to the task, feature, and label. Similar to the T-table template, it organizes the tabular data in a markdown format. 

\input{KDD2024_Camera/app_templates/table_example}

\section{Implementation Details}
\label{app:imp_details}

\subsection{Baseline Models}
\label{app:imp_details_baseline}

Our experiments encompass two categories of baselines: large language models (LLMs) and traditional tabular models. For LLMs, we employ instruction following to predict the answer in both zero-shot and in-context scenarios. In the case of tabular models, we only compare results in few-shot settings. Specifically, for learning-based models such as XGBoost~\citep{chen2016XGBoost}, CatBoost~\citep{prokhorenkova2018catboost}, LightGBM~\citep{ke2017LightGBM}, FTT~\citep{gorishniy2021revisit_tab_dnn}, and logistic regression (LR), we utilize in-context examples for training. Conversely, for TabPFN~\citep{hollmann2022TabPFN} and our GTL model, no parameter tuning is required; direct inference is achieved using the same examples through in-context learning. In order to conduct a fair comparison for all these baselines, we would sample category-balanced in-context samples for classification tasks. 

\subsubsection{LLMs Baseline}
We assess the instruction following capabilities of several large language models, dividing them into two categories: black box models and white box models. Black box models include the GPT series, where we can only access the response rather than model parameters and prediction logits. In contrast, white box models like LLaMA, allow us to access the entire model and obtain logits for each class. The descriptions of each baseline model are as follows:

\begin{itemize}
\item \textbf{GPT-3.5}~\citep{Brown2020GPT-3}: We employ the GPT-3.5-turbo version, which is the most capable GPT-3.5 model optimized for chat and designed for diverse tasks such as natural language understanding, translation, and summarization. To encourage the model to provide predictions and minimize instances where it refuses to make predictions, we incorporate an additional answer instruction prompt when querying GPT-3.5. This approach allows GPT-3.5 to predict the probabilities for each category, which can then be used to calculate AUROC.
\item \textbf{GPT-4}~\citep{OpenAI2023GPT4TR}: A powerful baseline renowned for its top performance in numerous language tasks due to its larger model size and architectural refinements. We also fine-tune the answer instruction prompt for GPT-4. Compared to GPT-3.5, GPT-4 demonstrates superior performance in both zero-shot and in-context learning. The instruction prompt and the response format is the same with GPT-3.5.
\item \textbf{LLaMA}~\citep{touvron2023llama2}: We employ the LLaMA-2 7B and 13B version as our baseline and backbone for generative tabular learning.
\end{itemize}

\subsubsection{Tradition Tabular Models Baseline}
For traditional tabular models, to optimize the results for each model, we have utilized and finely tuned common pre-processing methods. These methods include z-score normalization for numerical features and labels, and one-hot encoding for categorical features. The performance of these baseline models, following the application of these pre-processing strategies, are included in our results.

\begin{itemize}
\item \textbf{Logistic Regression}: A linear model used to predict binary response probabilities based on predictor variables. Both Logistic Regression and its regression counterpart, Linear Regression, are straightforward and efficient, often serving as baseline models in classification and regression tasks respectively. To enhance model performance, we employ z-score normalization on numerical features and one-hot encoding on categorical features.
\item \textbf{XGBoost}~\citep{chen2016XGBoost}: An advanced gradient-boosted decision tree algorithm renowned for its efficiency and high performance. Popular for handling diverse datasets and commonly used in machine learning competitions.
\item \textbf{LightGBM}~\citep{ke2017LightGBM}: A gradient boosting framework employing tree-based learning algorithms, designed for efficiency and scalability, offering faster training speeds and lower memory usage than other techniques.
\item \textbf{CatBoost}\citep{prokhorenkova2018catboost}: A high-performance gradient boosting library that handles categorical features directly, providing an efficient and accurate model. We implemented z-score normalization on numerical features and regression task labels.
\item \textbf{FTTransformers}\citep{gorishniy2021revisit_tab_dnn}: This model employs transformer-based architectures for tabular data, offering a dynamic and adaptable framework that can be fine-tuned for various tasks. In these deep learning-based models, we've found that the results largely depend on pre-processing strategies and the number of training epochs. Therefore, we selected the optimal strategy based on results from the holdout datasets. We implemented z-score normalization on numerical features and regression task labels. 
\item \textbf{TabPFN}~\citep{hollmann2022TabPFN} A deep learning model that combines Positional Feature-wise Networks with transformer-based architectures for tabular data, capturing both local and global feature interactions to enhance performance in various tasks.
\end{itemize}

\subsection{Generative Tabular Learning (GTL)}
\label{app:imp_details_GTL}

\paragraph{Hyper-parameters.}
We employ the LLaMA-2~\citep{touvron2023llama2} 7B and 13B version models as the backbone for our experiments. For generative tabular learning, we utilize a fixed learning rate of 1e-5 and a batch size of 512, without incorporating any scheduler or warmup. The training procedure involves gradient updates with the optimizer AdamW~\citep{loshchilov2018decoupled}. We set the limitation of maximum token numbers to 4096, which ensures that all samples are within the acceptable range and prevents truncation.

\paragraph{Experimental Environments.}
The GTL-enhanced LLaMA model is implemented using PyTorch version 2.1.0 and executed on CUDA 12.1, running on NVIDIA Tesla A100 GPUs. As for GTL, pre-training is performed on a single node equipped with 8 A100 GPUs and the micro batch size is 4 for LLaMA-7B and 2 for LLaMA-13B. With an overall batch size of 512, the gradient accumulation steps amount to 16 for LLaMA-7B and 32 for LLaMA-13B. In LLaMA-13B-GTL, updating the gradient for 256 batches, which comprises 131,072 samples, takes approximately 26 hours.

%% file: KDD2024_Camera/app_templates/table_example.tex
\begin{figure*}[!tb]
\centering
\begin{lstlisting}[basicstyle=\small, caption=Example of the T-table template with in-context examples, label=lst:t_table_cont_eg]
@You are an expert in medical diagnosis.@
@Based on the features of the patient, please predict the diabetes diagnosis.@
I will supply multiple instances with features and the corresponding label for reference.
Please refer to the table below for detailed descriptions of the features and label:
--- feature description ---
Age: The age of the individual in years
BMI: Body Mass Index of the individual
FBS: Fasting blood sugar level of the individual
HbA1c: Hemoglobin A1c level, a test to measure blood sugar level over the past 2 to 3 months
Gender: The gender of the individual, Male or Female
Blood Pressure: Blood pressure level of the individual, can be Normal or High
Family History: Whether the individual has a family history of diabetes or not
Smoking: Whether the individual smokes or not
Diet: The quality of the individual diet, can be Healthy or Poor
Exercise: Whether the individual exercises regularly or not
--- label description --- 
Diagnosis: Whether the individual is diagnosed with diabetes or not
--- data ---
|Age|BMI|FBS|HbA1c|Gender|Blood Pressure|Family History|Smoking|Diet|Exercise|Diagnosis|
*|48|47.0|200.0|9.2|Male|High|Yes|Yes|Poor|No|0|
|70|35.0|140.0|7.1|Female|Normal|No|No|Healthy|Regular|0|
|12|10.0|80.0|5.0|Male|Low|No|Yes|Poor|No|1|
|75|40.0|160.0|7.8|Female|High|Yes|Yes|Poor|No|1|*
|30|20.0|100.0|5.7|Female|Normal|No|No|Healthy|Regular|<MASK>|
@Please use the supplied data to predict the <MASK> Diagnosis. Diagnosed with diabetes[1] or not[0]?@
Answer: 0
\end{lstlisting}
\end{figure*}